\documentclass[runningheads]{llncs}
\usepackage{graphicx}

\usepackage{comment}
\usepackage{amsmath,amssymb} 
\usepackage{color}
\usepackage{algorithm}
\usepackage{algorithmic}

\usepackage{hyperref}
\hypersetup{
    colorlinks=false,
    pdfborder={0 0 0},
}
\usepackage{breqn}
\usepackage{enumitem}
\usepackage{multirow}
\usepackage[symbol]{footmisc}

\begin{document}
\pagestyle{headings}
\mainmatter
\def\ECCVSubNumber{2590}  

\title{Learning Joint Spatial-Temporal Transformations for Video Inpainting}

\titlerunning{Spatial-Temporal Transformer Networks for Video Inpainting}
%
\author{Yanhong Zeng\inst{1,2*} \and
Jianlong Fu\inst{3}$^\dagger$ \and
Hongyang Chao\inst{1,2}$^\dagger$}
\authorrunning{Y. Zeng, J. Fu, and H. Chao}
%
\institute{School of Data and Computer Science, Sun Yat-sen University, Guangzhou, China  \and 
Key Laboratory of Machine Intelligence and Advanced Computing, Ministry of Education, China \and
Microsoft Research Asia\\
\email{zengyh7@mail2.sysu.edu.cn}, \email{isschhy@mail.sysu.edu.cn}, \email{jianf@microsoft.com}}
\maketitle 

\begin{abstract}
   \footnotetext[1]{This work was done when Y. Zeng was an intern at Microsoft Research Asia.}
   \footnotetext[2]{ J. Fu and H. Chao are the corresponding authors.}

High-quality video inpainting that completes missing regions in video frames is a promising yet challenging task. State-of-the-art approaches adopt attention models to complete a frame by searching missing contents from reference frames, and further complete whole videos frame by frame. 
However, these approaches can suffer from inconsistent attention results along spatial and temporal dimensions, which often leads to blurriness and temporal artifacts in videos. 
In this paper, we propose to learn a joint \textbf{S}patial-\textbf{T}emporal \textbf{T}ransformer \textbf{N}etwork (\textbf{STTN}) for video inpainting. 
Specifically, we simultaneously fill missing regions in all input frames by self-attention, and propose to optimize STTN by a spatial-temporal adversarial loss. 
To show the superiority of the proposed model, we conduct both quantitative and qualitative evaluations by using standard stationary masks and more realistic moving object masks. Demo videos are available at \url{https://github.com/researchmm/STTN}.

\end{abstract}

\section{Introduction}
\label{sec:intro}
Video inpainting is a task that aims at filling missing regions in video frames with plausible contents \cite{bertalmio2001navier}. An effective video inpainting algorithm has a wide range of practical applications, such as corrupted video restoration \cite{granados2012not}, unwanted object removal \cite{matsushita2006full,patwardhan2005video}, video retargeting \cite{kim2019deep} and under/over-exposed image restoration \cite{lee2019copy}. 
Despite of the huge benefits of this technology, high-quality video inpainting still meets grand challenges, such as the lack of high-level understanding of videos \cite{kim2019deepblind,wang2019video} and high computational complexity \cite{chang2019free,xu2019deep}. 

\begin{figure}
   \begin{center}
      \includegraphics[width=\linewidth]{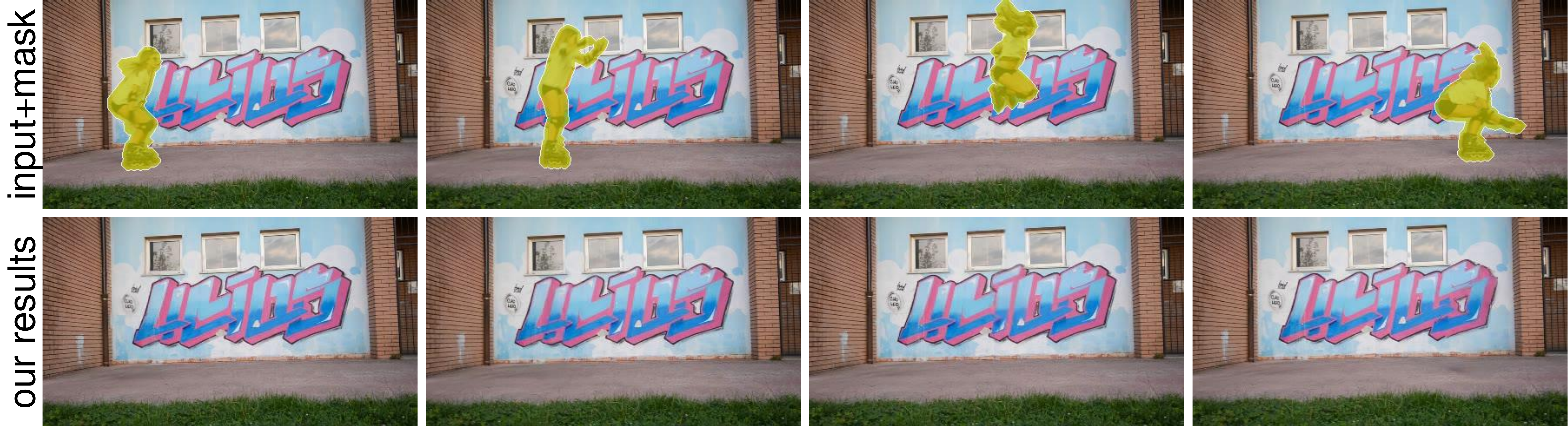}
   \end{center}
   \caption{We propose \textbf{S}patial-\textbf{T}emporal \textbf{T}ransformer \textbf{N}etworks for completing missing regions in videos in a spatially and temporally coherent manner. The top row shows sample frames with yellow masks denoting user-selected regions to be removed. The bottom row shows our completion results. [Best viewed with zoom-in]}
   \label{fig:teaser}
\end{figure}

Significant progress has been made by using 3D convolutions and recurrent networks for video inpainting \cite{chang2019free,kim2019deep,wang2019video}. These approaches usually fill missing regions by aggregating information from nearby frames. However, they suffer from temporal artifacts due to limited temporal receptive fields.
To solve the above challenge, state-of-the-art methods apply attention modules to capture long-range correspondences, so that visible contents from distant frames can be used to fill missing regions in a target frame \cite{lee2019copy,oh2019onion}. 
One of these approaches synthesizes missing contents by a weighting sum over the aligned frames with frame-wise attention \cite{lee2019copy}. The other approach proposes a step-by-step fashion, which gradually fills missing regions with similar pixels from boundary towards the inside by pixel-wise attention \cite{oh2019onion}. Although promising results have been shown, these methods have two major limitations due to the significant appearance changes caused by complex motions in videos.
One limitation is that these methods usually assume global affine transformations or homogeneous motions, 
which makes them hard to model complex motions and often leads to inconsistent matching in each frame or in each step. 
Another limitation is that all videos are processed frame by frame without specially-designed optimizations for temporal coherence. Although post-processing is usually used to stabilize generated videos, it is usually time-costing. Moreover, the post-processing may fail in cases with heavy artifacts.

To relieve the above limitations, we propose to learn a joint \textbf{S}patial-\textbf{T}emporal \textbf{T}ransformer \textbf{N}etwork (\textbf{STTN}) for video inpainting.
We formulate video inpainting as a ``multi-to-multi'' problem, which takes both neighboring and distant frames as input and simultaneously fills missing regions in all input frames.
To fill missing regions in each frame, 
the transformer searches coherent contents from all the frames along both spatial and temporal dimensions by a proposed multi-scale patch-based attention module. 
Specifically, patches of different scales are extracted from all the frames to cover different appearance changes caused by complex motions. 
Different heads of the transformer calculate similarities on spatial patches across different scales.
Through such a design, the most relevant patches can be detected and transformed for the missing regions by aggregating attention results from different heads. 
Moreover, the spatial-temporal transformers can be fully exploited by stacking multiple layers, so that attention results for missing regions can be improved based on updated region features.
Last but not least, we further leverage a spatial-temporal adversarial loss for joint optimization \cite{chang2019free,chang2019learnable}. 
Such a loss design can optimize STTN to learn both perceptually pleasing and coherent visual contents for video inpainting. 

In summary, our main contribution is to learn joint spatial and temporal transformations for video inpainting, by a deep generative model with adversarial training along spatial-temporal dimensions. 
Furthermore, the proposed multi-scale patch-based video frame representations can enable fast training and inference, which is important to video understanding tasks.
We conduct both quantitative and qualitative evaluations using both stationary masks and moving object masks for simulating real-world applications (e.g., watermark removal and object removal). 
Experiments show that our model outperforms the state-of-the-arts by a significant margin in terms of PSNR and VFID with relative improvements of 2.4\% and 19.7\%, respectively.
We also show extensive ablation studies to verify the effectiveness of the proposed spatial-temporal transformer.

\section{Related Work}
\label{sec:relate}
To develop high-quality video inpainting technology, many efforts have been made on filling missing regions with spatially and temporally coherent contents in videos \cite{bertalmio2001navier,huang2016temporally,lee2019copy,newson2014video,wang2019video,xu2019deep}. 
We discuss representative patch-based methods and deep generative models for video inpainting as below.

\textbf{Patch-based methods:} 
Early video inpainting methods mainly formulate the inpainting process as a patch-based optimization problem \cite{barnes2009patchmatch,criminisi2004region,patwardhan2005video,wexler2007space}. 
Specifically, these methods synthesize missing contents by sampling similar spatial or spatial-temporal patches from known regions based on a global optimization \cite{newson2014video,patwardhan2007video,wexler2007space}. Some approaches try to improve performance by providing foreground and background segments \cite{granados2012not,patwardhan2005video}. Other works focus on joint estimations for both appearance and optical-flow \cite{huang2016temporally,matsushita2006full}. 
Although promising results can be achieved, patch-based optimization algorithms typically assume a homogeneous motion field in holes and they are often limited by complex motion in general situations. 
Moreover, optimization-based inpainting methods often suffer from high computational complexity, which is infeasible for real-time applications \cite{xu2019deep}.

\textbf{Deep generative models:} 
With the development of deep generative models, significant progress has been made by deep video inpainting models. Wang et al. are the first to propose to combine 3D and 2D fully convolution networks for learning temporal information and spatial details for video inpainting \cite{wang2019video}. However, the results are blurry in complex scenes. Xu et al. improve the performance by jointly estimating both appearance and optical-flow \cite{xu2019deep,zhang2019internal}. Kim et al. adopt recurrent networks for ensuring temporal coherence \cite{kim2019deep}. Chang et al. develop Temporal SN-PatchGAN \cite{yu2019free} and temporal shift modules \cite{lin2019tsm} for free-form video inpainting \cite{chang2019free}. 
Although these methods can aggregate information from nearby frames, they fail to capture visible contents from distant frames. 

To effectively model long-range correspondences, recent models have adopted attention modules and show promising results in image and video synthesis \cite{ma2018gan,yang2020learning,zeng2019learning}.
Specifically, Lee et al. propose to synthesize missing contents by weighted summing aligned frames with frame-wise attention \cite{lee2019copy}. However, the frame-wise attention relies on global affine transformations between frames, which is hard to handle complex motions. 
Oh et al. gradually fill holes step by step with pixel-wise attention \cite{oh2019onion}. Despite promising results, it is hard to ensure consistent attention result in each recursion. Moreover, existing deep video inpainting models that adopt attention modules process videos frame by frame without specially-designed optimization for ensuring temporal coherence.

\section{Spatial-Temporal Transformer Networks}
\label{sec:app}
\subsection{Overall design}
\label{sec:over}

\begin{figure}[t]
   \begin{center}
      \includegraphics[width=\linewidth]{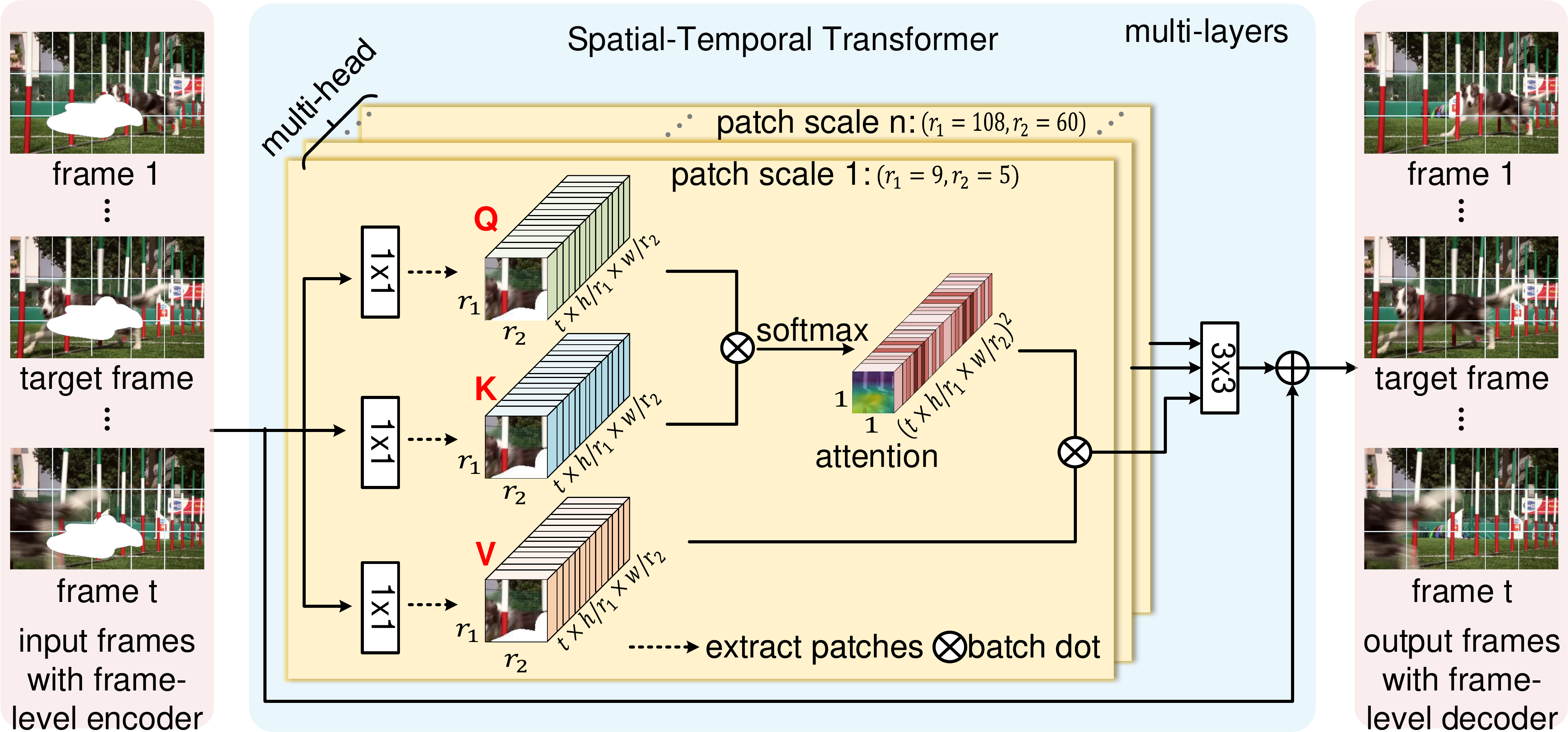}
   \end{center}
   \caption{\textbf{Overview of the Spatial-Temporal Transformer Networks (STTN).}
   STTN consists of 1) a frame-level encoder, 2) multi-layer multi-head spatial-temporal transformers and 3) a frame-level decoder. The transformers are designed to simultaneously fill holes in all input frames with coherent contents. Specifically, a transformer matches the queries (Q) and keys (K) on spatial patches across different scales in multiple heads, thus the values (V) of relevant regions can be detected and transformed for the holes. Moreover, the transformers can be fully exploited by stacking multiple layers to improve attention results based on updated region features. $1\times 1$ and $3\times 3$ denote the kernel size of 2D convolutions. More details can be found in Section \ref{sec:app}. }
   \label{fig:sttn}
\end{figure}

\textbf{Problem formulation:}
Let $X_1^T := \{X_1, X_2, ..., X_T \}$ be a corrupted video sequence of height $H$, width $W$ and frames length $T$.
$M_1^T := \{M_1, M_2, ..., M_T\}$ denotes the corresponding frame-wise masks. For each mask $M_i$, value ``0'' indicates known pixels, and value ``1'' indicates missing regions. 
We formulate deep video inpainting as a self-supervised task that randomly creates ($X_1^T, M_1^T$) pairs as input and reconstruct the original video frames $Y_1^T = \{Y_1, Y_2, ..., Y_T\}$. 
Specifically, we propose to learn a mapping function from masked video $X_1^T$ to the output $\hat{Y}_1^T := \{\hat{Y}_1, \hat{Y}_2, ..., \hat{Y}_T\}$, such that the conditional distribution of the real data  $p(Y_1^T|X_1^T)$ can be approximated by the one of generated data $p(\hat{Y}_1^T | X_1^T)$. 

The intuition is that an occluded region in a current frame would probably be revealed in a region from a distant frame, especially when a mask is large or moving slowly. 
To fill missing regions in a target frame, it is more effective to borrow useful contents from the whole video by taking both neighboring frames and distant frames as conditions.
To simultaneously complete all the input frames in a single feed-forward process, we formulate the video inpainting task as a ``multi-to-multi'' problem. Based on the Markov assumption \cite{hausman1999independence}, we simplify the ``multi-to-multi'' problem and denote it as:
\begin{equation}
   p(\hat{Y}_1^T|X_1^T) = \prod_{t=1}^{T} p(\hat{Y}_{t-n}^{t+n}| X_{t-n}^{t+n}, X_{1, s}^T), 
   \label{eq1}
\end{equation} 
where $X_{t-n}^{t+n}$ denotes a short clip of neighboring frames with a center moment $t$ and a temporal radius $n$. $X_{1, s}^T$ denotes distant frames that are uniformly sampled from the videos $X_1^T$ in a sampling rate of $s$. 
Since $X_{1, s}^T$ can usually cover most key frames of the video, it is able to describe ``the whole story'' of the video. Under this formulation, video inpainting models are required to not only preserve temporal consistency in neighboring frames, but also make the completed frames to be coherent with ``the whole story'' of the video.

\textbf{Network design:}
The overview of the proposed \textbf{S}patial-\textbf{T}emporal \textbf{T}ransfo-rmer \textbf{N}etworks (\textbf{STTN}) is shown in Figure \ref{fig:sttn}. 
As indicated in Eq. (\ref{eq1}), STTN takes both neighboring frames $X^{t+n}_{t-n}$ and distant frames $X^T_{1,s}$ as conditions, and complete all the input frames simultaneously.
Specifically, STTN consists of three components, including a frame-level encoder, multi-layer multi-head spatial-temporal transformers, and a frame-level decoder. The frame-level encoder is built by stacking several 2D convolution layers with strides, which aims at encoding deep features from low-level pixels for each frame. Similarly, the frame-level decoder is designed to decode features back to frames. 
Spatial-temporal transformers are the core component, which aims at learning joint spatial-temporal transformations for all missing regions in the deep encoding space.

\subsection{Spatial-temporal transformer}
\label{sec:sttn}

To fill missing regions in each frame, spatial-temporal transformers are designed to search coherent contents from all the input frames. 
Specifically, we propose to search by a multi-head patch-based attention module along both spatial and temporal dimensions.
Different heads of a transformer calculate attentions on spatial patches across different scales. Such a design allows us to handle appearance changes caused by complex motions.
For example, on one hand, attentions for patches of large sizes (e.g., frame size $H\times W$) aim at completing stationary backgrounds. On the other hand, attentions for patches of small sizes (e.g., $\frac{H}{10}\times \frac{W}{10}$) encourage capturing deep correspondences in any locations of videos for moving foregrounds.

A multi-head transformer runs multiple ``Embedding-Matching-Attending'' steps for different patch sizes in parallel. In the Embedding step, features of each frame are mapped into query and memory (i.e., key-value pair) for further retrieval. In the Matching step, region affinities are calculated by matching queries and keys among spatial patches that are extracted from all the frames. Finally, relevant regions are detected and transformed for missing regions in each frame in the Attending step. We introduce more details of each step as below. 

\textbf{Embedding:} 
We use $f_1^T = \{f_1, f_2, ..., f_T\}$, where $f_i \in R ^{h\times w\times c}$ to denote the features encoded from the frame-level encoder or former transformers, which is the input of transformers in Fig. \ref{fig:sttn}. 
Similar to many sequence modeling models, mapping features into key and memory embeddings is an important step in transformers \cite{girdhar2019video,vaswani2017attention}. Such a step enables modeling deep correspondences for each region in different semantic spaces:
\begin{equation}
   q_i, (k_i, v_i) = M_q(f_i), (M_k(f_i), M_v(f_i)),
\end{equation}
where $1 \leq i \leq T$, $M_q(\cdot)$, $M_k(\cdot)$ and $M_v(\cdot)$ denote the $1\times 1$ 2D convolutions that embed input features into query and memory (i.e., key-value pair) feature spaces while maintaining the spatial size of features.

\textbf{Matching:} 
We conduct patch-based matching in each head. In practice, we first extract spatial patches of shape $r_1 \times r_2 \times c$ from the query feature of each frame, and we obtain $N = T\times h/r_1 \times w/r_2$ patches. Similar operations are conducted to extract patches in the memory (i.e., key-value pair in the transformer). 
Such an effective multi-scale patch-based video frame representation can avoid redundant patch matching and enable fast training and inference.
Specifically, we reshape the query patches and key patches into 1-dimension vectors separately, so that patch-wise similarities can be calculated by matrix multiplication. The similarity between $i$-th patch and $j$-th patch is denoted as:
\begin{equation}\label{eq:sim}
   s_{i,j} = \frac{{\boldsymbol{p^q_i}}\cdot (\boldsymbol{p^k_j})^T}{\sqrt{r_1\times r_2\times c}}, 
\end{equation}
where $1 \leq i, j \leq N$, $\boldsymbol{p^q_i}$ denotes the $i$-th query patch, $\boldsymbol{p^k_j}$ denotes the $j$-th key patch. The similarity value is normalized by the dimension of each vector to avoid a small gradient caused by subsequent softmax function \cite{vaswani2017attention}. 
Corresponding attention weights for all patches are calculated by a softmax function:
\begin{equation}\label{eq:soft}
\alpha_{i, j} = 
\left\{\begin{matrix} 
 \exp(s_{i,j}) / \sum\limits_{\substack{n=1}}^N \exp(s_{i, n}), & \boldsymbol{p_j} \in \Omega, & \\ 
  & & \\
   0, &  \boldsymbol{p_j} \in \bar{\Omega}. & 
\end{matrix}\right.
\end{equation} 
where $\Omega$ denotes visible regions outside masks, and $\bar{\Omega}$ denotes missing regions. Naturally, we only borrow features from visible regions for filling holes.

\textbf{Attending:} 
After modeling the deep correspondences for all spatial patches, the output for the query of each patch can be obtained by weighted summation of values from relevant patches:
\begin{equation}
   o_i = \sum_{j=1}^{N} \alpha_{i,j} \boldsymbol{p_j^v},
\end{equation} 
where $\boldsymbol{p_j^v}$ denotes the $j$-th value patch. After receiving the output for all patches, we piece all patches together and reshape them into $T$ frames with original spatial size $h\times w\times c$. The resultant features from different heads are concatenated and further passed through a subsequent 2D residual block \cite{he2016deep}. This subsequent processing is used to enhance the attention results by looking at the context within the frame itself.

\begin{figure}[t]
   \begin{center}
      \includegraphics[width=\linewidth]{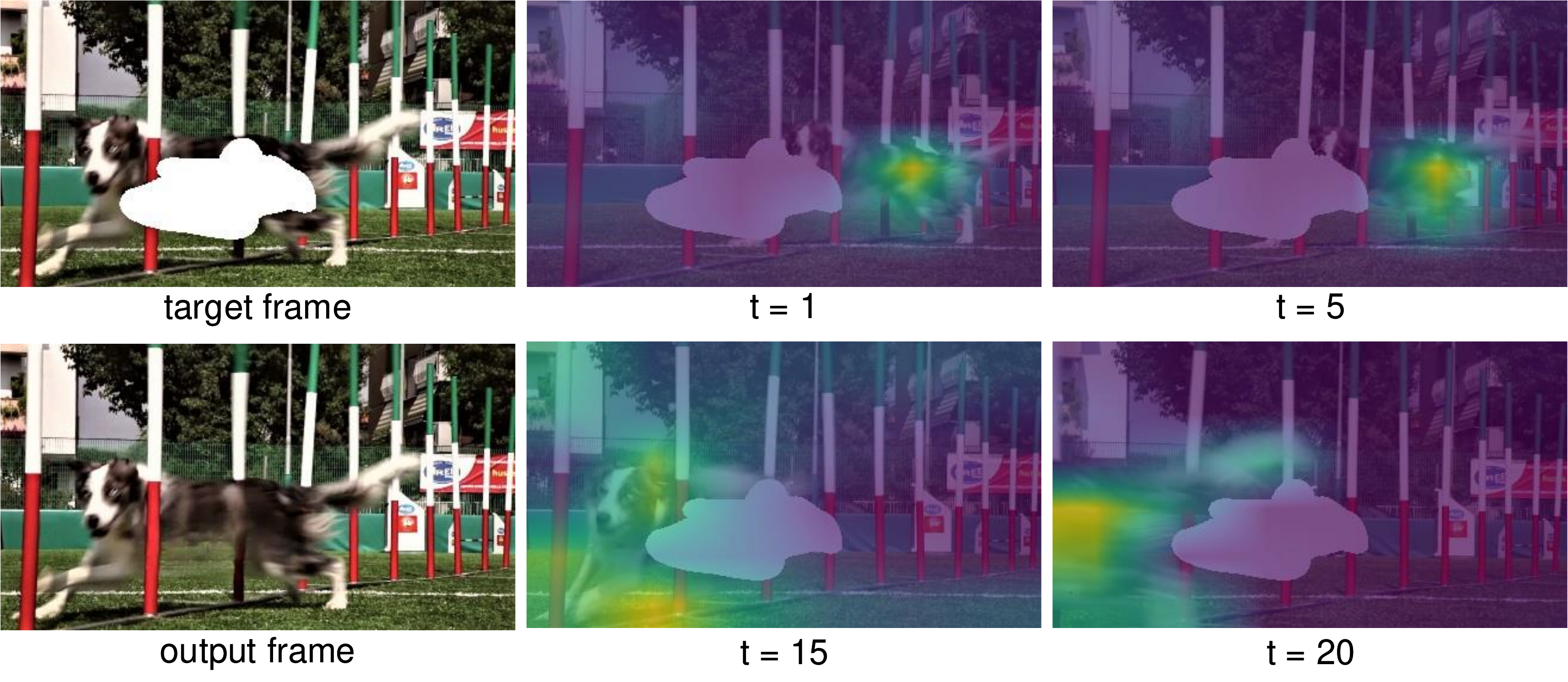}
   \end{center}
   \caption{\textbf{Illustration of the attention maps for missing regions learned by STTN.} For completing the dog corrupted by a random mask in a target frame (e.g., t=10), our model is able to ``track'' the moving dog over the video in both spatial and temporal dimensions. Attention regions are highlighted in bright yellow.}
   \label{fig:attn-dog}
\end{figure}

The power of the proposed transformer can be fully exploited by stacking multiple layers, so that attention results for missing regions can be improved based on updated region features in a single feed-forward process. Such a multi-layer design promotes learning coherent spatial-temporal transformations for filling in missing regions. 
As shown in Fig. \ref{fig:attn-dog}, we highlight the attention maps learned by STTN in the last layer in bright yellow. 
For the dog partially occluded by a random mask in a target frame, spatial-temporal transformers are able to ``track'' the moving dog over the video in both spatial and temporal dimensions and fill missing regions in the dog with coherent contents.

\subsection{Optimization objectives}
\label{sec:opt}
As outlined in Section \ref{sec:over}, we optimize the proposed STTN in an end-to-end manner by taking the original video frames as ground truths without any other labels. The principle of choosing optimization objectives is to ensure per-pixel reconstruction accuracy, perceptual rationality and spatial-temporal coherence in generated videos \cite{chang2019free,gatys2016image,johnson2016perceptual,lee2019copy}. To this end, we select a pixel-wise reconstruction loss and a spatial-temporal adversarial loss as our optimization objectives.

In particular, we include $L_1$ losses calculated between generated frames and original frames for ensuring per-pixel reconstruction accuracy in results. The $L_1$ losses for hole regions are denoted as:
\begin{equation}
   L_{hole} =  \frac{\|M_1^T \odot (Y_1^T - \hat{Y}_1^T)\|_1}{\|M_1^T\|_1} ,
\end{equation}
and corresponding $L_1$ losses for valid regions are denoted as:
\begin{equation}
   L_{valid} =  \frac{\| (1-M_1^T) \odot (Y_1^T - \hat{Y}_1^T)\|_1}{\|1-M_1^T\|_1}, 
\end{equation}
where $\odot$ indicates element-wise multiplication, and the values are normalized by the size of corresponding regions. 

Inspired by the recent studies that adversarial training can help to ensure high-quality content generation results, we propose to use a Temporal PatchGAN (T-PatchGAN) as our discriminator \cite{chang2019free,chang2019learnable,yang2020learning,zeng2019learning}. 
Such an adversarial loss has shown promising results in enhancing both perceptual quality and spatial-temporal coherence in video inpainting \cite{chang2019free,chang2019learnable}. 
In particular, the T-PatchGAN is composed of six layers of 3D convolution layers. The T-PatchGAN learns to distinguish each spatial-temporal feature as real or fake, so that spatial-temporal coherence and local-global perceptual details of real data can be modeled by STTN. 
The detailed optimization function for the T-PatchGAN discriminator is shown as follows: 
\begin{equation}
   L_D = E_{x\sim P_{Y^T_1}(x)} [ReLU(1-D(x))] + E_{z\sim P_{\hat{Y}^T_1}(z)}[ReLU(1+D(z))],
\end{equation} 
and the adversarial loss for STTN is denoted as:
\begin{equation}
   L_{adv} = -E_{z\sim P_{\hat{Y}^T_1}(z)} [D(z)].
\end{equation}
The overall optimization objectives are concluded as below:
\begin{equation}
   L = \lambda_{hole} \cdot L_{hole} + \lambda_{valid} \cdot L_{valid}  + \lambda_{adv} \cdot L_{adv}.
\end{equation}  

We empirically set the weights for different losses as: $\lambda_{hole} = 1$, $L_{valid} = 1$, $L_{adv}=0.01$. Since our model simultaneously complete all the input frames in a single feed-forward process, our model runs at 24.3 fps on a single GPU NVIDIA V100. More details are provided in the Section D of our supplementary material.

\section{Experiments}
\label{sec:exp}

\subsection{Dataset} 
\label{subsec:data}

To evaluate the proposed model and make fair comparisons with SOTA approaches, we adopt the two most commonly-used datasets in video inpainting, including Youtube-VOS \cite{xu2018youtube} and DAVIS \cite{caelles20182018}. 
In particular, \textbf{YouTube-VOS} contains 4,453 videos with various scenes, including bedrooms, streets, and so on. The average video length in Youtube-VOS is about 150 frames. We follow the original train/validation/test split (i.e., 3,471/474/508) and report experimental results on the test set for Youtube-VOS. 
In addition, we also evaluate different approaches on 
\textbf{DAVIS} dataset \cite{caelles20182018}, as this dataset is composed of 150 high-quality videos of challenging camera motions and foreground motions. We follow the setting in previous works \cite{kim2019deep,xu2019deep}, and set the training/testing split as 60/90 videos. Since the training set of DAVIS is limited (60 videos with at most 90 frames for each), we initialize model weights by a pre-trained model on YouTube-VOS following the settings used in \cite{kim2019deep,xu2019deep}.

To simulate real-world applications, we evaluate models by using two types of free-form masks, including stationary masks and moving masks \cite{chang2019learnable,kim2019deep,lee2019copy}. Because free-form masks are closer to real masks and have been proved to be effective for training and evaluating inpainting models \cite{chang2019free,chang2019learnable,liu2018image,nazeri2019edgeconnect}. 
Specifically, for testing \textbf{stationary masks}, we generate stationary random shapes as testing masks to simulate applications like watermark removal. More details of the generation algorithm are provided in the Section B of our supplementary material.
Since this type of application targets at reconstructing original videos, we take original videos as ground truths and evaluate models from both quantitative and qualitative aspects. 
For testing \textbf{moving masks}, we use foreground object annotations as testing masks to simulate applications like object removal. Since the ground truths after foreground removal are unavailable, we evaluate the models through qualitative analysis following previous works \cite{kim2019deep,lee2019copy,xu2019deep}.

\subsection{Baselines and evaluation metrics}
\label{subsec:base}
Recent deep video inpainting approaches have shown state-of-the-art performance with fast computational time \cite{kim2019deep,lee2019copy,oh2019onion,xu2019deep}. To evaluate our model and make fair comparisons, we select the most recent and the most competitive approaches for comparisons, which are listed as below:
\begin{itemize}[nosep]
   \item \textbf{VINet \cite{kim2019deep}} adopts a recurrent network to aggregate temporal features from neighboring frames. 
   \item \textbf{DFVI \cite{xu2019deep}} fills missing regions in videos by pixel propagation algorithm based on completed optical flows.
   \item \textbf{LGTSM \cite{chang2019learnable}} proposes a learnable temporal shift module and a spatial-temporal adversarial loss for ensuring spatial and temporal coherence. 
   \item \textbf{CAP \cite{lee2019copy}} synthesizes missing contents by a deep alignment network and a frame-based attention module. 
\end{itemize}
We fine-tune baselines multiple times on YouTube-VOS \cite{xu2018youtube} and DAVIS \cite{caelles20182018} by their released models and codes and report their best results in this paper.

We report quantitative results by four numeric metrics, i.e., PSNR \cite{xu2019deep}, SSIM \cite{chang2019free}, flow warping error \cite{lai2018learning} and video-based Fr\'{e}chet Inception Distance (VFID) \cite{chang2019free,wang2018video}. Specifically, we use PSNR and SSIM as they are the most widely-used metrics for video quality assessment. 
Besides, the flow warping error is included to measure the temporal stability of generated videos. 
Moreover, FID has been proved to be an effective perceptual metric and it has been used by many inpainting models \cite{oh2019onion,wang2018video,zhang2018unreasonable}. In practice, we use an I3D \cite{carreira2017quo} pre-trained video recognition model to calculate VFID following the settings in \cite{chang2019free,wang2018video}. 

\subsection{Comparisons with state-of-the-arts}
\label{subsec:comp}

\textbf{Quantitative Evaluation:}
We report quantitative results for filling stationary masks on Youtube-VOS \cite{xu2018youtube} and DAVIS \cite{caelles20182018} in Table \ref{tb:exp}. 
As stationary masks often involve partially occluded foreground objects, it is challenging to reconstruct a video especially with complex appearances and object motions.
Table \ref{tb:exp} shows that, compared with SOTA models, our model performs better video reconstruction quality with both per-pixel and overall perceptual measurements. 
Specifically, our model outperforms the SOTA models by a significant margin, especially in terms of PSNR, flow warp error and VFID. 
The specific gains are 2.4\%, 1.3\% and 19.7\% relative improvements on Youtube-VOS, respectively. The superior results show the effectiveness of the proposed spatial-temporal transformer and adversarial optimizations in STTN.

\begin{table}[!htbp]
\begin{center}
\begin{tabular}{r|r||c|c||c|c}
   \multicolumn{2}{r||}{Models} &PSNR$^\star$ &SSIM ($\%$)$^\star$ &$E_{warp}$ ($\%$)$^\dag$ &VFID$^\dag$ \\\hline\hline
  \multirow{5}{*}{\rotatebox{90}{Youtube-vos}}  
  &VINet\cite{kim2019deep} &29.20 &94.34  &0.1490 &0.072  \\\cline{2-6}
  &DFVI \cite{xu2019deep}  &29.16 &94.29  &0.1509 &0.066 \\\cline{2-6}
  &LGTSM \cite{chang2019learnable} &29.74  &95.04 &0.1859 &0.070 \\\cline{2-6}
  &CAP \cite{lee2019copy} &31.58 &96.07 &0.1470  &0.071  \\\cline{2-6}
  ~&Ours &\textbf{32.34}  &\textbf{96.55}  &\textbf{0.1451} &\textbf{0.053}  \\\hline\hline
  
  \multirow{5}{*}{\rotatebox{90}{DAVIS}}  
  &VINet\cite{kim2019deep} &28.96 &94.11 &0.1785 &0.199  \\\cline{2-6}
  &DFVI \cite{xu2019deep}          &28.81 &94.04 &0.1880 &0.187 \\\cline{2-6}
  &LGTSM \cite{chang2019learnable} &28.57 &94.09 &0.2566 &0.170  \\\cline{2-6}
  &CAP \cite{lee2019copy}          &30.28 &95.21 &0.1824 &0.182  \\\cline{2-6}
  ~&Ours &\textbf{30.67} &\textbf{95.60} &\textbf{0.1779} &\textbf{0.149} \\\hline
\end{tabular}
\end{center}
\caption{Quantitative comparisons with state-of-the-art models on Youtube-VOS \cite{xu2018youtube} and DAVIS \cite{caelles20182018}. Our model outperforms baselines in terms of PSNR \cite{xu2019deep}, SSIM \cite{chang2019free}, flow warping error ($E_{warp}$) \cite{lai2018learning} and VFID \cite{wang2018video}. $^\star$ Higher is better. $^\dag$ Lower is better. }
\label{tb:exp}
\end{table}

\begin{figure}[t]
   \begin{center}
      \includegraphics[width=\linewidth]{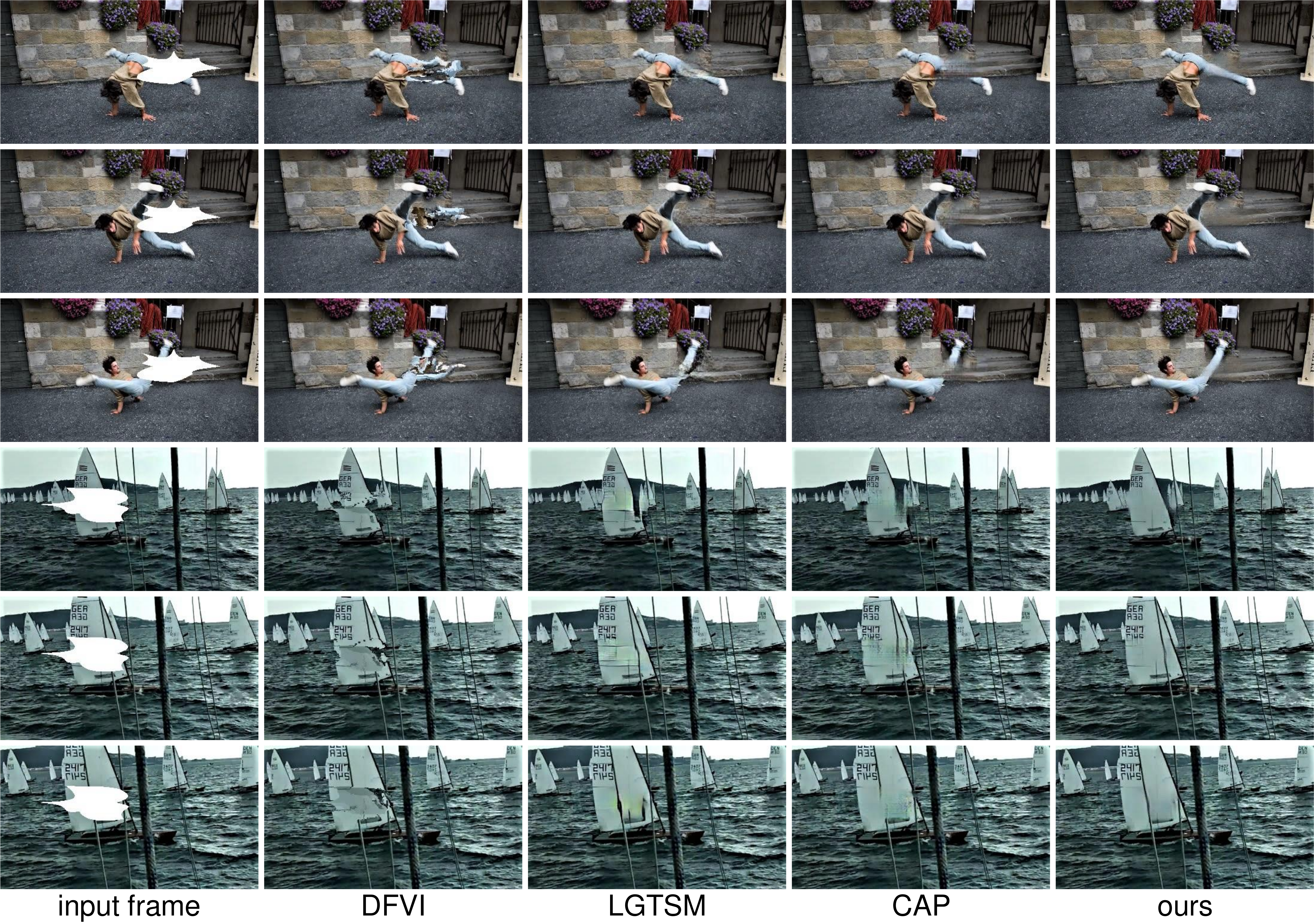}
   \end{center}
   \caption{\textbf{Visual results for stationary masks}. The first column shows input frames from DAVIS \cite{caelles20182018} (top-3) and YouTube-VOS \cite{xu2018youtube} (bottom-3), followed by results from DFVI \cite{xu2019deep}, LGTSM \cite{chang2019learnable}, CAP \cite{lee2019copy}, and our model. Comparing with the SOTAs, our model generates more coherent structures and details of the legs and boats in results.}
   \label{fig:visfix}
\end{figure}
\begin{figure}
   \begin{center}
      \includegraphics[width=\linewidth]{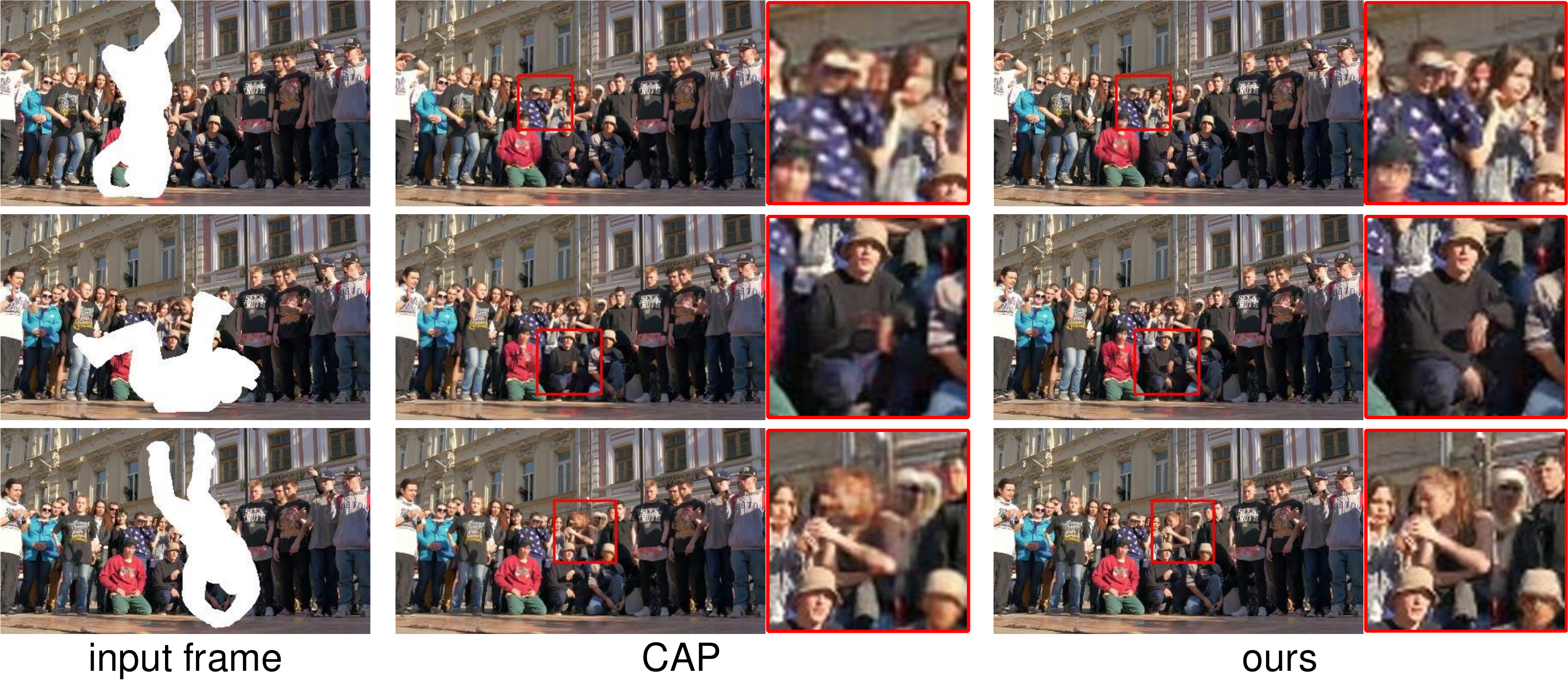}
   \end{center}
   \caption{\textbf{Visual comparisons for filling moving masks.} Comparing with CAP \cite{lee2019copy}, one of the most competitive models for filling moving masks, our model is able to generate visually pleasing results even under complex scenes (e.g., clear faces for the first and the third frames, and better results than CAP for the second frame). }
   \label{fig:visobj}
\end{figure}

\textbf{Qualitative Evaluation:}
For each video from test sets, we take all frames for testing. To compare visual results from different models, we follow the setting used by most video inpainting works and randomly sample three frames from the video for case study \cite{lee2019copy,oh2019onion,wang2019video}. 
We select the most three competitive models, DFVI \cite{xu2019deep}, LGTSM \cite{chang2019learnable} and CAP \cite{lee2019copy} for comparing results for stationary masks in Fig. \ref{fig:visfix}. 
We also show a case for filling in moving masks in Fig. \ref{fig:visobj}. 
To conduct pair-wise comparisons and analysis in Fig. \ref{fig:visobj}, we select the most competitive model, CAP \cite{lee2019copy}, according to the quantitative comparison results.
We can find from the visual results that our model is able to generate perceptually pleasing and coherent contents in results. 
More video cases are available online\footnote{video demo: \url{https://github.com/researchmm/STTN}}.

\begin{figure}
   \begin{center}
      \includegraphics[width=\linewidth]{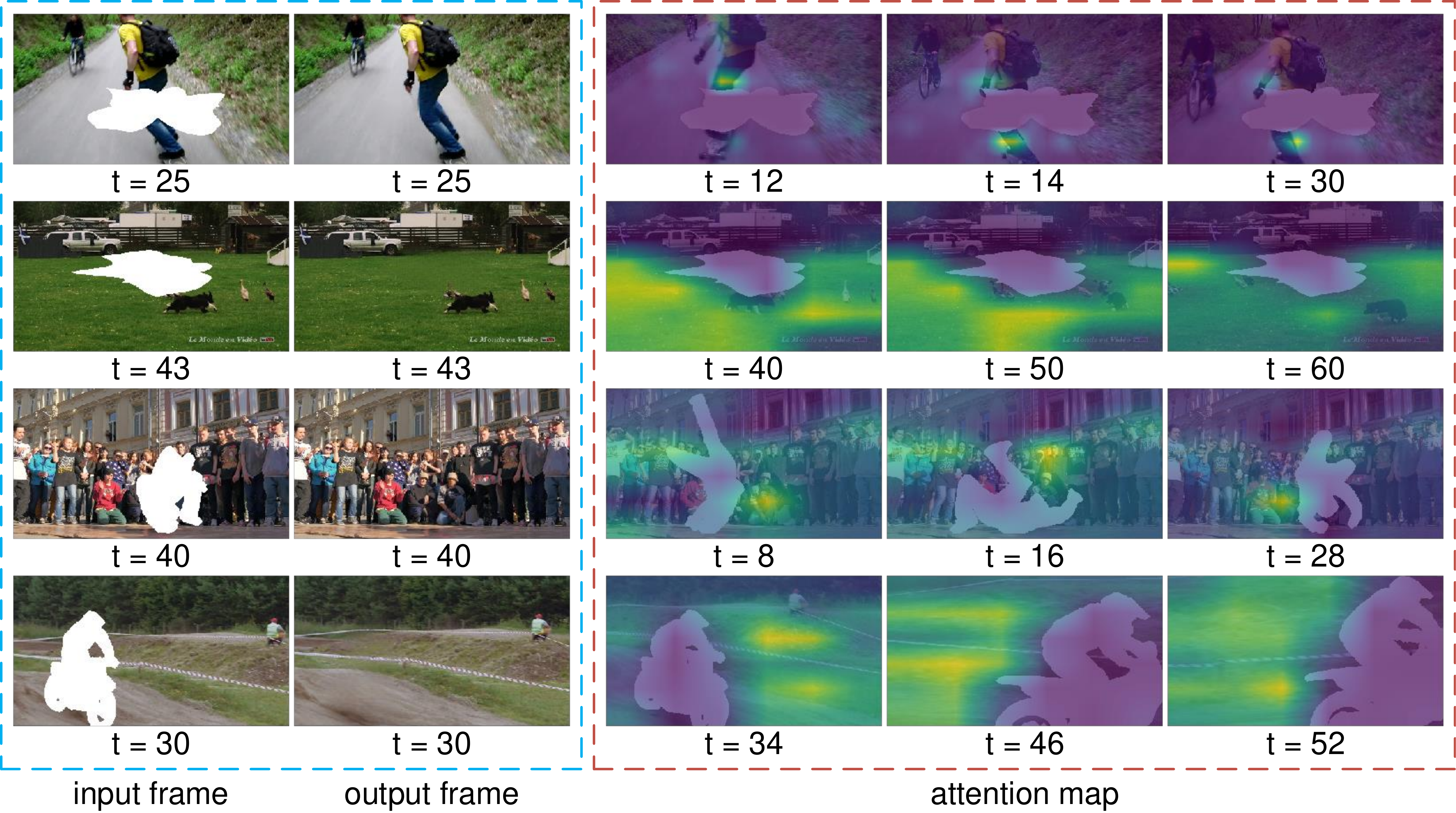}
   \end{center}
   \caption{
  \textbf{Illustration of attention maps for missing regions learned by the proposed STTN.}
   We highlight the most relevant patches in yellow according to attention weights. For filling partially occluded objects (the first and the third cases), STTN can precisely attend to the objects. For filling backgrounds (the second and the fourth cases), STTN can correctly attend to the backgrounds.}
   \label{fig:visatn}
\end{figure}

In addition to visual comparisons, we visualize the attention maps learned by STTN in Fig. \ref{fig:visatn}. Specifically, we highlight the top three relevant regions captured by the last transformer in STTN in bright yellow. 
The relevant regions are selected according to the attention weights calculated by Eq. (\ref{eq:soft}). We can find in Fig. \ref{fig:visatn} that STTN is able to precisely attend to the objects for filling partially occluded objects in the first and the third cases. For filling the backgrounds in the second and the fourth cases, STTN can correctly attend to the backgrounds.

\begin{figure}
   \begin{center}
      \includegraphics[width=0.9\linewidth]{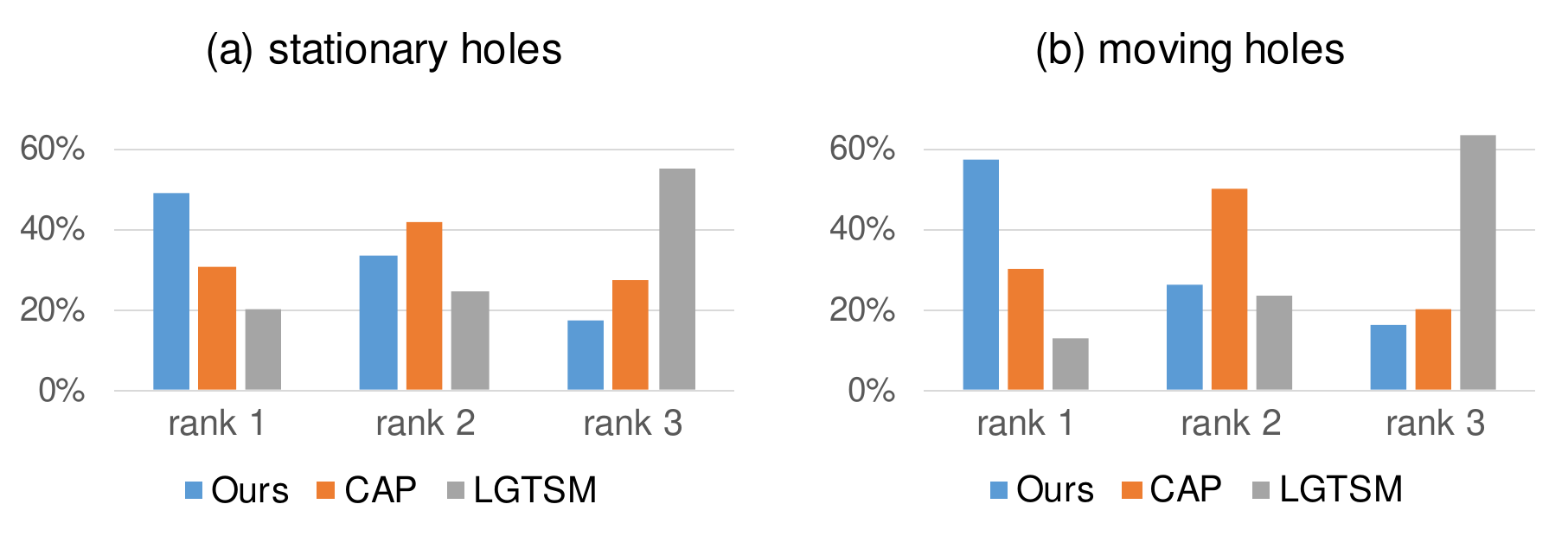}
   \end{center}
   \caption{User study. ``Rank x'' means the percentage of results from each model being chosen as the x-th best. Our model is ranked in first place in most cases.}
   \label{fig:user}
\end{figure}

\textbf{User Study:} 
We conduct a user study for a more comprehensive comparison. 
we choose LGTSM \cite{chang2019learnable} and CAP \cite{lee2019copy} as two strong baselines, since we have observed their significantly better performance than other baselines from both quantitative and qualitative results. 
We randomly sampled 10 videos (5 from DAVIS and 5 from YouTube-VOS) for stationary masks filling, and 10 videos from DAVIS for moving masks filling. In practice, 28 volunteers are invited to the user study. In each trial, inpainting results from different models are shown to the volunteers, and the volunteers are required to rank the inpainting results. 
To ensure a reliable subjective evaluation, videos can be replayed multiple times by volunteers. Each participant is required to finish 20 groups of trials without time limit. 
Most participants can finish the task within 30 minutes. 
The results of the user study are concluded in Fig \ref{fig:user}. We can find that our model performs better in most cases for these two types of masks. 

\subsection{Ablation Study}
\label{subsec:ab}
To verify the effectiveness of the spatial-temporal transformers, this section presents ablation studies on DAVIS dataset \cite{caelles20182018} with stationary masks. 
More ablation studies can be found in the Section E of our supplementary material.

\textbf{Effectiveness of multi-scale:} 
To verify the effectiveness of using multi-scale patches in multiple heads, we compare our model with several single-head STTNs with different patch sizes. 
In practice, we select patch sizes according to the spatial size of features, so that the features can be divided into patches without overlapping. 
The spatial size of features in our experiments is $108\times 60$.
Results in Table \ref{tb:ab-size} show that our full model with multi-scale patch-based video frame representation achieves the best performance under this setting. 

\begin{table}
   \begin{center}
   \begin{tabular}{r|c|c|c|c} 
      Patch size &PSNR$^\star$ &SSIM($\%$)$^\star$ &$E_{warp}$ ($\%$)$^\dag$ &VFID$^\dag$ \\\hline \hline
    $108\times 60$  &30.16  &95.16 &0.2243 &0.168 \\\hline 
    $36 \times 20$  &30.11  &95.13 &0.2051 &0.160 \\\hline 
    $18 \times 10$  &30.17  &95.20 &0.1961 &0.159 \\\hline 
    $9 \times 5$    &30.43  &95.39 &0.1808 &0.163 \\\hline 
    Ours &\textbf{30.67} &\textbf{95.60} &\textbf{0.1779} &\textbf{0.149} \\\hline
   \end{tabular} 
\end{center}
   \caption{Ablation study by using different patch scales in attention layers. Ours combines the above four scales. $^\star$ Higher is better. $^\dag$ Lower is better.}
   \label{tb:ab-size}
\end{table}

\textbf{Effectiveness of multi-layer:}
The spatial-temporal transformers can be stacked by multiple layers to repeat the inpainting process based on updated region features. We verify the effectiveness of using multi-layer spatial-temporal transformers in Table \ref{tb:ab-stack}. 
We find that stacking more transformers can bring continuous improvements and the best results can be achieved by stacking eight layers. Therefore, we use eight layers in transformers as our full model.

\begin{table}
   \begin{center}
   \begin{tabular}{r|c|c|c|c} 
     Stack  &PSNR$^\star$ &SSIM($\%$)$^\star$ &$E_{warp}$ ($\%$)$^\dag$ &VFID$^\dag$ \\\hline \hline
     $\times 2$   &30.17 &95.17 &0.1843 &0.162 \\\hline 
    $\times 4$   &30.38 &95.37 &0.1802 &0.159 \\\hline 
    $\times 6$  &30.53 &95.47 &0.1797 &0.155 \\\hline 
    $\times 8$ (ours) &\textbf{30.67} &\textbf{95.60} &\textbf{0.1779} &\textbf{0.149} \\\hline 
   \end{tabular} 
\end{center}
   \caption{Ablation study by using different stacking number of the proposed spatial-temporal transformers. $^\star$ Higher is better. $^\dag$ Lower is better.}
   \label{tb:ab-stack}
\end{table}


\begin{figure}
   \begin{center}
      \includegraphics[width=\linewidth]{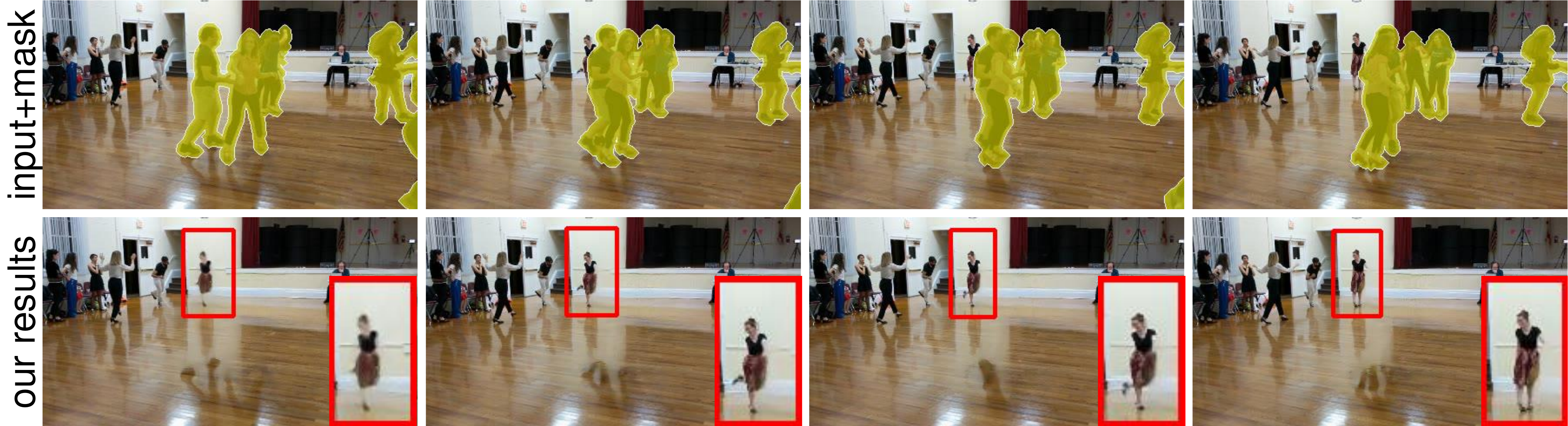}
   \end{center}
   \caption{A failure case. The bottom row shows our results with enlarged patches in the bottom right corner. For reconstructing the dancing woman occluded by a large mask, STTN fails to generate continuous motions and it generates blurs inside the mask.}
   \label{fig:failure}
\end{figure}

\section{Conclusions}
In this paper, we propose a novel joint spatial-temporal transformation learning for video inpainting. Extensive experiments have shown the effectiveness of multi-scale patch-based video frame representation in deep video inpainting models. Coupled with a spatial-temporal adversarial loss, our model can be optimized to simultaneously complete all the input frames in an efficient way. The results on YouTube-VOS \cite{xu2018youtube} and DAVIS \cite{caelles20182018} with challenging free-form masks show the state-of-the-art performance by our model.

We note that STTN may generate blurs in large missing masks if continuous quick motions occur. As shown in Fig. \ref{fig:failure}, STTN fails to generate continuous dancing motions and it generates blurs when reconstructing the dancing woman in the first frame. We infer that STTN only calculates attention among spatial patches, and the short-term temporal continuity of complex motions are hard to capture without 3D representations. 
In the future, we plan to extend the proposed transformer by using attention on 3D spatial-temporal patches to improve the short-term coherence. We also plan to investigate other types of temporal losses \cite{lai2018learning,wang2018video} for joint optimization in the future.

\section*{Acknowledgments}
This project was supported by NSF of China under Grant 61672548, U1611461.

\clearpage
\bibliographystyle{splncs04}
\bibliography{egbib}

\clearpage

\section*{Supplementary Material}
This supplementary material presents the details of complete video inpainting results in Section \ref{sec:vid} and our stationary mask generation algorithm in Section \ref{sec:mask}. 
We provide the details of our network architectures in Section \ref{sec:net} and the implementation details in Section \ref{sec:imp}. 
Finally, extensive ablation studies and analysis for the proposed Spatial-Temporal Transformer Networks for video inpainting can be found in Section \ref{sec:ab}.

\setcounter{section}{0}
\def\thesection{\Alph{section}}
\section{Video Inpainting Results}
\label{sec:vid}

To compare visual results from different inpainting models in our main paper, we follow the setting used in most video inpainting works \cite{huang2016temporally,kim2019deep,xu2019deep}. Specifically, we sample several frames from video results and show them in Figure 4 and Figure 5 in the main paper. 
However, sampled frames cannot truly reflect video results. Sometimes sampled static frames look less blurry but artifacts can be stronger in a dynamic video. 
Therefore, we provide 20 video cases for a more comprehensive comparison\footnote{video demo: \url{https://github.com/researchmm/STTN}}.

In practice, we test all the videos in the test sets of DAVIS dataset \cite{caelles20182018} (90 cases) and Youtube-VOS dataset \cite{xu2018youtube} (508 cases), and we randomly show 20 cases for visual comparisons. 
Specifically, five cases from DAVIS and five cases from Youtube-VOS are used to test filling stationary masks. Since Youtube-VOS has no dense object annotations, we sample 10 videos with dense object annotations from DAVIS to test filling moving masks following the setting used in previous works \cite{kim2019deep,lee2019copy,xu2019deep}. 
To conduct side-by-side comparisons and analysis, we select the two most competitive video inpainting models, LGTSM \cite{chang2019learnable} and CAP \cite{lee2019copy} in the videos. LGTSM and CAP are fine-tuned multiple times to achieve optimal video results by the codes and models publicly provided by their official Github homepage\footnote{
LGTSM: \url{https://github.com/amjltc295/Free-Form-Video-Inpainting}\\
CAP: \url{https://github.com/shleecs/Copy-and-Paste-Networks-for-Deep-Video-Inpainting}
}. We can find from the video results that our model outperforms the state-of-the-art models in most cases.

\section{Stationary Mask Generation Algorithm}
\label{sec:mask}

Inspired by Xu et al. \cite{xu2019deep}, we use stationary masks and moving masks as testing masks to simulate real-world applications (e.g., watermark removal and object removal) in the main paper. 
As introduced in Section 4.1 in the main paper, on one hand, we use frame-wise foreground object annotations from DAVIS datasets \cite{caelles20182018} as moving masks to simulate applications like object removal. 
On the other hand, we generate random shapes as stationary masks to simulate applications like watermark removal. Specifically, for the task of removing watermarks, a user often draw a mask along the outline of a watermark. 
Inspired by previous mask generation algorithms \cite{chang2019free,yu2019free}, we propose a stationary mask generation algorithm to simulate such a behavior for drawing masks for watermarks. 
Specifically, the proposed algorithm randomly generates a set of control points around a unit circle, and then it smoothly connects these points into a closed cyclic contour by cubic Bezier curves. The details of the stationary mask generation algorithm are shown in Algorithm \ref{al:1} as follows.

\begin{algorithm}
   \caption{Algorithm for stationary mask generation. \textsl{maxPointNum}, \textsl{maxLength} are hyper-parameters to control the statinary mask generation.}
   \label{al:1}
   \begin{algorithmic}
      \STATE mask = zeros(imgHeight, imgWidth)
      \STATE pointNum = random.uniform(maxPointNum)
      \STATE startX = origX = random.uniform(imgWidth)
      \STATE startY = origY = random.uniform(imgHeight)
      \STATE angles = linspace(0, 2*pi, pointNum)
      \FOR{i=0 to pointNum} 
      \STATE length = random.uniform(maxLength)
      \STATE x = sin(angles[i]) * length
      \STATE y = cos(angles[i]) * length 
      \STATE // comment: ensuring smoothness of contours
      \STATE Connect (startX, startY) to (x, y) by cubic Bezier curves. 
      \STATE startX = x 
      \STATE startY = y 
      \ENDFOR
      \STATE // comment: ensuring a closed cyclic contour
      \STATE Connect (startX, startY) to (origX, origY) by cubic Bezier curves. 
   \end{algorithmic}
\end{algorithm}

\section{Details of Network Architecture}
\label{sec:net}

The Spatial-Temporal Transformer Network (STTN) is built upon a generative adversarial framework. Specifically, the proposed STTN plays a role as a generator in the framework, and we adopt a Temporal PatchGAN (T-PatchGAN) \cite{chang2019free} as our discriminator. 
The T-PatchGAN is composed of six layers of 3D convolution layers. Specifically, the T-PatchGAN learns to classify each spatial-temporal feature as real or fake, while STTN learns to fool the T-PatchGAN. Such an adversarial training allows STTN to model the local-global perceptual rationality and the spatial-temporal coherence of real videos \cite{chang2019free}. 
In addition to the introduction in Section 3 in the main paper, we provide the details of the architectures of STTN and the T-PatchGAN in Table \ref{tb:gen} and Table \ref{tb:dis}, respectively. 
Specifically, features inside holes are computed by dilated 2D convolutions. We argue that STTN is able to leverages multi-scale contexts and updates holes' features multiple times to improve attention results. 

\begin{table}
\begin{center}
\begin{tabular}{c|c|c|c|c} 
Module Name &Filter Size &\# Channels &Stride/Up Factor &Nonlinearity \\\hline\hline
2dConv &$3\times3$ &64 &2 &LeakyReLU(0.2) \\ 
2dConv &$3\times3$ &64 &1 &LeakyReLU(0.2) \\ 
2dConv &$3\times3$ &128 &2 &LeakyReLU(0.2) \\ 
2dConv &$3\times3$ &256 &1 &LeakyReLU(0.2) \\ \hline
\multirow{2}{*}{Transformer $\times$ 8}   &$1\times1$ &\multirow{2}{*}{256} &1  &- \\
~ &$3\times3$ & &1 &LeakyReLU(0.2) \\\hline
BilinearUpSample &- &256 &2 &-    \\
2dConv &$3\times3$ &128 &1 &LeakyReLU(0.2) \\ 
2dConv &$3\times3$ &64 &1 &LeakyReLU(0.2) \\ \hline
BilinearUpSample &- &64 &2 &-    \\
2dConv &$3\times3$ &64 &1 &LeakyReLU(0.2) \\ 
2dConv &$3\times3$ &3 &1 &Tanh \\ \hline
\end{tabular} 
 \end{center}
    \caption{Details of the proposed Spatial-Temporal Transformer Networks (STTN). ``2dConv'' means 2D convolution layers. ``Transformer $\times$ 8'' denotes stacking the proposed spatial-temporal transformers by eight layers. A transformer layer involves $1\times1$ and $3\times3$ convolutions (The overview of STTN is shown in Fig. 2 in the main paper). We use bilinear interpolations for all upsample operations on feature maps \cite{liu2018image,oh2019onion}. We show whether and what nonlinearity layer is used in the nonlinearity column.}
    \label{tb:gen}
 \end{table}

 \begin{table}
    \begin{center}
    \begin{tabular}{c|c|c|c|c} 
    Module Name &Filter Size &\# Channels &Stride &Nonlinearity \\\hline\hline
    SN-3dConv &$3\times5\times5$ &64 &(1,2,2) &LeakyReLU(0.2) \\ 
    SN-3dConv &$3\times5\times5$ &128 &(1,2,2) &LeakyReLU(0.2) \\ 
    SN-3dConv &$3\times5\times5$ &256 &(1,2,2) &LeakyReLU(0.2) \\ 
    SN-3dConv &$3\times5\times5$ &256 &(1,2,2) &LeakyReLU(0.2) \\ 
    SN-3dConv &$3\times5\times5$ &256 &(1,2,2) &LeakyReLU(0.2) \\ 
    SN-3dConv &$3\times5\times5$ &256 &(1,2,2) &-\\ \hline
    \end{tabular} 
     \end{center}
    \caption{Details of the Temporal-PatchGAN (T-PatchGAN) discriminator \cite{chang2019free}. The T-PatchGAN is composed of six 3D convolution layers. ``SN-3dConv'' denotes a 3D convolution layer that adopts spectral normalization to stabilize GAN's training \cite{chang2019free}.}
    \label{tb:dis}
    \end{table}

\section{Implementation details}
\label{sec:imp}

\textbf{Hyper-parameters:}
To maintain the aspect ratio of videos and take into account the memory limitations of modern GPUs, we resize all video frames into $432 \times 240$ for both training and testing \cite{huang2016temporally,kim2019deep,lee2019copy,xu2019deep}. During training, we set the batch size as 8, and the learning rate starts with 1e-4 and decays with factor 0.1 every 150k iterations. Specifically, for each iteration, we sample five frames from a video in a consecutive or discontinuous manner with equal probability for training following Lee et al. \cite{lee2019copy,oh2019onion}.

\textbf{Computation complexity:} Our full model has a total of 12.6M trainable parameters. It costs about 3.9G GPU memory for completing a video from DAVIS dataset \cite{caelles20182018} by STTN on average. 
The proposed multi-scale patch-based video frame representations can enable fast training and inference. Specifically, our model runs at about 24.3fps with an NVIDIA V100 GPU and it runs at about 10.43 fps with an NVIDIA P100 GPU on average. Its total training time was about 3 days on YouTube-VOS dataset \cite{xu2018youtube} and one day for fine-tuning on DAVIS dataset \cite{caelles20182018} with 8 Tesla V100 GPUs. 
The computation complexity of the proposed spatial-temporal transformers are denoted as: 
\begin{equation}
   \mathcal{O}(\sum_{l=1}^{D}\left [ 2\cdot (n\cdot \frac{HW}{p_wp_h})^2 \cdot (p_wp_hC_l) + nk^2_lHWC_{l-1}C_l \right ]) \approx \mathcal{O}(n^2),
   \label{eq:comp}
\end{equation}
where $D$ is the number of transformer layers, $n$ is the number of input frames, $HW$ is the feature size, $p_wp_h$ is the patch size, $k_l$ denotes for kernel size, and $C$ is the channel number of features. In Eq. (\ref{eq:comp}), we focus on the computation complexity caused by the spatial-temporal transformers and leave out other computation costs (e.g., encoding and decoding costs) for simplification.

\section{More ablation studies}
\label{sec:ab}

To verify the effectiveness of the proposed Spatial-Temporal Transformer Networks (STTN) for video inpainting, this section presents extensive ablation studies on DAVIS dataset \cite{caelles20182018} with stationary masks. 

\textbf{Effectiveness of utilizing distant frames:}
we test our full model with different sample rates to prove the benefits of utilizing distant frames. Quantitative comparison results on DAVIS dataset \cite{caelles20182018} with stationary masks can be found in Table \ref{tb:ab-frames}. The first row ($s>T$) means that the STTN takes only neighboring frames as input. Besides, the second row ($s=20$) means that the STTN takes both neighboring frames and distant frames that are uniformly sampled from the videos in a sampling rate of 20 frames. 

Table \ref{tb:ab-frames} shows that leveraging visible contexts in distant frames helps in generating better results especially in terms of VFID with 5.70\% relative improvements. Based on the observation that most videos in YouTube-VOS dataset \cite{xu2018youtube} and DAVIS dataset \cite{caelles20182018} won't vary a lot within 10 frames on average, we set the sample rate as 10 in our full model to avoid sampling redundant frames and to save computation costs. 

\begin{table}
   \begin{center}
   \begin{tabular}{r|c|c|c|c} 
    Sample Rate &PSNR$^\star$ &SSIM($\%$)$^\star$ &$E_{warp}$ ($\%$)$^\dag$ &VFID$^\dag$ \\\hline \hline
    $s > T$ &30.55 &95.47 &0.1802 &0.158 \\\hline 
    $s=20$  &30.62 &95.55 &0.1790 &0.152 \\\hline 
    $s=10$ (ours) &\textbf{30.67} &\textbf{95.60} &\textbf{0.1779} &\textbf{0.149} \\\hline 
   \end{tabular} 
\end{center}
   \caption{Ablation study by utilizing distant frames in different sampling rates. Our full model set $s=10$. $^\star$ Higher is better. $^\dag$ Lower is better. }
   \label{tb:ab-frames}
\end{table}

\textbf{Effectiveness of masked normalization:} 
As shown in Eq. (3) and Eq. (4) in the main paper, we normalize the value of similarity by the dimension of vectors and filter out unknown regions for similarities calculating. 
In this part, we conduct comparisons between models with or without such a masked normalization in Table \ref{tb:ab-norm}. Results show that such an operation is necessary since it brings improvements with a significant margin comparing with the one without masked normalization.  
\begin{table}
   \begin{center}
   \begin{tabular}{r|c|c|c|c} 
    &PSNR$^\star$ &SSIM($\%$)$^\star$ &$E_{warp}$ ($\%$)$^\dag$ &VFID$^\dag$ \\\hline \hline
    w/o masked norm.  &30.39 &95.32 &0.1849 &0.162 \\\hline 
    w/ masked norm. &\textbf{30.67} &\textbf{95.60} &\textbf{0.1779} &\textbf{0.149} \\\hline 
   \end{tabular} 
\end{center}
   \caption{Ablation study for the effectiveness of masked normalization operation on similarity calculation. $^\star$ Higher is better. $^\dag$ Lower is better. }
   \label{tb:ab-norm}
\end{table}


\textbf{Effectiveness of the Temporal PatchGAN Loss:}
Recent state-of-the-art deep video inpainting models that adopt attention modules often include a perceptual loss \cite{johnson2016perceptual}  and a style loss \cite{gatys2016image} as optimization objectives for perceptually pleasing results \cite{lee2019copy,oh2019onion}. However, they do not leverage specially-designed losses for ensuring temporal coherence. 
Chang et al. propose a novel Temporal PatchGAN (T-PatchGAN) loss for ensuring both perceptual rationality and spatial-temporal coherence of videos \cite{chang2019free,chang2019learnable}. However, they only apply T-PatchGAN on consecutive frames while the attention-based deep video inpainting models take discontinuous frames as input for training. 
We are the first to introduce T-PatchGAN in video inpainting models that adopt attention modules and show that T-PatchGAN is also powerful in discontinuous frames. Such a joint optimization encourages STTN to learn both local-global perceptual rationality and coherent spatial-temporal transformations for video inpainting. 

\begin{figure}
   \begin{center}
      \includegraphics[width=\linewidth]{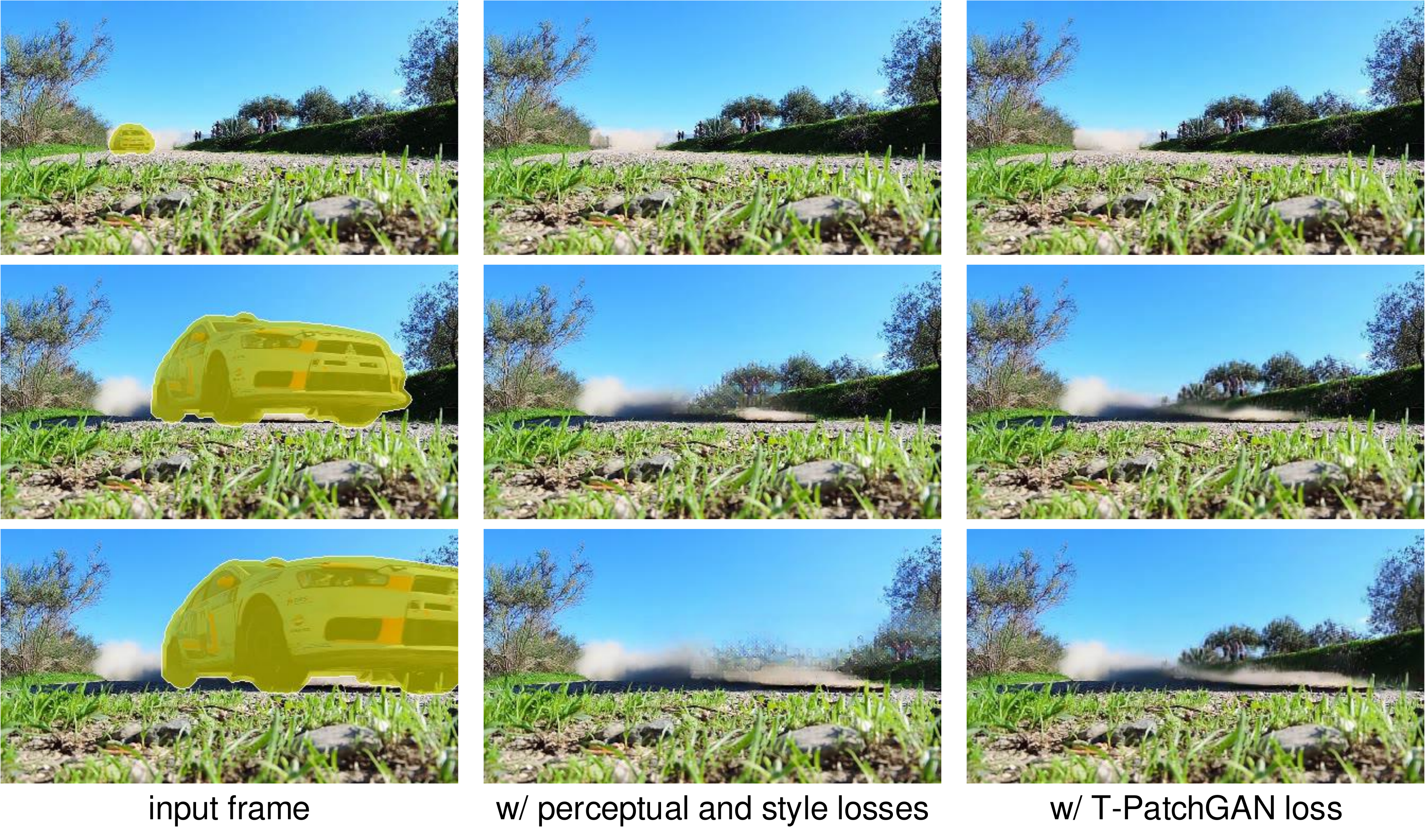}
   \end{center}
   \caption{Visual comparisons between an STTN optimized by a perceptual loss \cite{johnson2016perceptual} and a style loss \cite{gatys2016image} and an STTN optimized by a T-PatchGAN loss \cite{chang2019free}. These two models perform similarly in small missing regions, while in large missing regions, the model optimized by perceptual and style losses tends to generate artifacts in the missing regions. [Best viewed with zoom-in]}
   \label{fig:loss}
\end{figure}

We verify the effectiveness of the T-PatchGAN loss by quantitative comparisons in Table \ref{tb:ab-gan}. Compared with the STTN optimized by a style loss \cite{gatys2016image} and a perceptual loss \cite{johnson2016perceptual} following previous works \cite{lee2019copy,oh2019onion}, the STTN optimized by a T-PatchGAN loss performs better by a significant margin, especially in terms of VFID with 6.9\% relative improvements.
We also provide a visual comparison in Fig. \ref{fig:loss}. The visual results show that the STTN optimized by a T-PatchGAN loss can generate more coherent results than the one optimized by a perceptual loss and a style loss. The superior results show the effectiveness of the joint spatial-temporal adversarial learning in STTN. 

\begin{table}
   \begin{center}
   \begin{tabular}{r|c|c|c|c} 
    losses &PSNR$^\star$ &SSIM($\%$)$^\star$ &$E_{warp}$ ($\%$)$^\dag$ &VFID$^\dag$ \\\hline \hline
    w/ style \cite{gatys2016image}, w/ perceptual \cite{johnson2016perceptual} &30.38 &95.35 &0.1821 &0.160 \\\hline 
    w/ T-PatchGAN \cite{chang2019free} &\textbf{30.67} &\textbf{95.60} &\textbf{0.1779} &\textbf{0.149} \\\hline 
   \end{tabular} 
\end{center}
   \caption{Ablation study for different losses. $^\star$ Higher is better. $^\dag$ Lower is better. }
   \label{tb:ab-gan}
\end{table}

Specifically, perceptual loss and style loss have shown great impacts in many image generation tasks since they were proposed \cite{gatys2016image,johnson2016perceptual,liu2018image}. 
A perceptual loss computes $L_1$ distance between the activation maps of real frames and generated frames. A style loss is similar to the perceptual loss but aims at minimizing the $L_1$ distance between Gram matrices of the activation maps of real frames and generated frames. In practice, the activation maps are extracted from layers (e.g., \textsl{pool1}, \textsl{pool2} and \textsl{pool3}) of a pre-trained classification network (more details see \cite{lee2019copy,liu2018image,oh2019onion}). With the help of extracted low-level features, the perceptual loss and the style loss are helpful in generating high-frequency details. 

Unfortunately, perceptual and style losses are calculated on the features of a single frame and they are unable to leverage temporal contexts. When filling in a large missing region in videos, the perceptual and style losses are hard to enforce the generator to synthesize rational contents due to limited contexts. As a result, they have to generate meaningless high-frequency textures to match ground truths' low-level features. 
For example, for filling the large missing regions in the second and the third frames in Fig. \ref{fig:loss}, the STTN optimized by perceptual and style losses tends to generate high-frequency artifacts in the large missing regions. 
Similar artifacts can be found in the failure cases of previous works \cite{chang2019free,liu2018image}. Since the T-PatchGAN is able to leverage temporal contexts to optimize the generator, there are fewer artifacts in the results by using the T-PatchGAN.
For the above considerations, we use the T-PatchGAN loss instead of the perceptual and style losses in our final optimization objectives. In the future, we plan to design video-based perceptual and style losses which are computed on spatial-temporal features to leverage temporal contexts for optimization.

\end{document}